\documentclass[10pt,twocolumn,letterpaper]{article}

 \usepackage{wacv}              
\usepackage[accsupp]{axessibility}
\definecolor{wacvblue}{rgb}{0.21,0.49,0.74}
\usepackage[pagebackref,breaklinks,colorlinks,allcolors=wacvblue]{hyperref}

\usepackage{amsfonts}
\usepackage{xcolor}
\usepackage{bbding}
\usepackage{booktabs}
\usepackage{multirow}
\usepackage{balance}

\def\wacvPaperID{741} 
\def\confName{WACV}
\def\confYear{2027}

\newcommand{\maketitlesupp}{
\begin{center}
    {\Large \bf Unified Trajectory and 3D Pose Forecasting for Multi-Agent Sports Analysis\\
     --- Supplementary Material ---\par}
    \vspace*{8pt}
    {\large
     Marc Peral$^{1,2}$ \quad Guillem Capellera$^{1,2}$ \quad Luis Ferraz$^{2}$ \quad Antonio Rubio$^{2}$\\
     Stella Gra\ss hof$^{3}$ \quad Dan Witzner Hansen$^{3}$ \quad Antonio Agudo$^{1}$\\
     \vspace{1mm}
     \textcolor{gray}{$^{1}$Institut de Rob\`otica i Inform\`atica Industrial, CSIC-UPC}\\
     \textcolor{gray}{$^{2}$Kognia Sports Intelligence \quad $^{3}$IT University of Copenhagen}\par}
    \vspace*{8pt}
  \end{center}%
}

\title{Unified Trajectory and 3D Pose Forecasting for Multi-Agent Sports Analysis}

\author{Marc Peral$^{1,2}$ \quad Guillem Capellera$^{1,2}$ \quad Luis Ferraz$^{2}$ \quad Antonio Rubio$^{2}$ \\ Stella Graßhof$^{3}$ \quad Dan Witzner Hansen$^{3}$ \quad  Antonio Agudo$^{1}$ \\   \textcolor{gray}{$^1$Institut de Robòtica i Informàtica Industrial, CSIC-UPC} \\ \textcolor{gray}{$^2$Kognia Sports Intelligence \quad $^3$IT University of Copenhagen}}

\begin{document}
\maketitle
\begin{abstract}
Forecasting future human motion requires jointly reasoning about an agent's global trajectory and 3D body pose, which are tightly coupled, particularly in team sports where multiple interacting players introduce complex social dynamics. We exploit a socio-temporal encoder that fuses trajectory and pose at the input level and models them jointly through shared factorized attention. The model is further enhanced by a per-joint loss reweighting scheme that emphasizes dynamic joints and prevents degenerate solutions, and a frequency-domain trajectory smoothing term that improves trajectory prediction while indirectly benefiting pose forecasting. To ensure fair comparison with prior joint forecasting methods, we identify and correct evaluation biases arising from overlapping sequence construction, adopting instead a non-overlapping protocol and a deterministic, oracle-free strategy for multi-future prediction. Our approach improves over prior joint forecasting methods on both WorldPose and 3DPW, with the largest gains on pose accuracy, demonstrating that the proposed joint formulation generalizes beyond the densely interactive scenarios that motivate it.
\end{abstract}

\begin{figure}
\centering
  \includegraphics[width=0.85\linewidth]{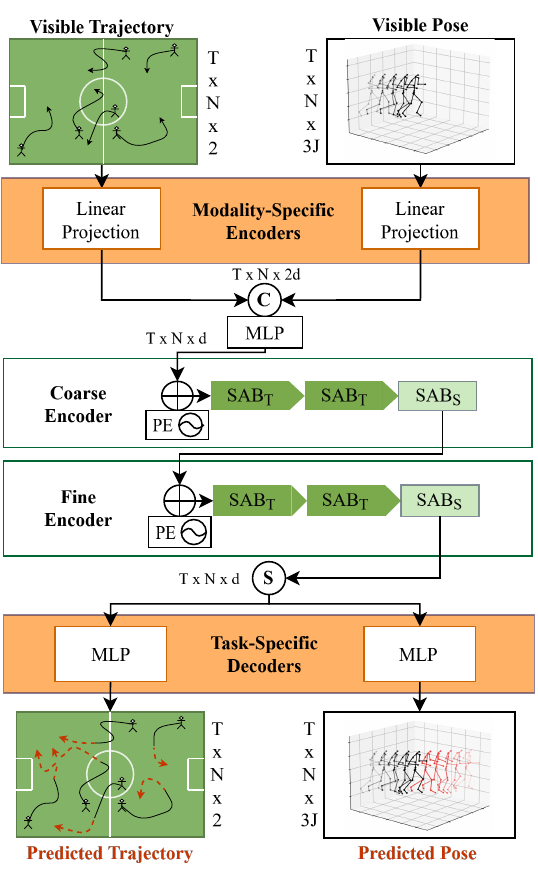}
  \vspace{-0.4cm}
  \caption{\textbf{Overview.} Prior trajectories and poses are first projected independently into a shared latent space. The fused representation is processed by a socio-temporal encoder whose output is fed into two task-specific heads implemented as independent MLPs.}
  \vspace{-0.4cm}
  \label{fig:overview}
\end{figure}

\section{Introduction}
\label{sec:intro}

Anticipating how a person will move and how their body will be configured a few seconds into the future is a problem with direct consequences for robotics, autonomous driving, surveillance, and sports analytics. Two largely separate lines of work have addressed pieces of this problem: one forecasts the trajectory that an agent will follow through space~\cite{aydemir2023adapt, choi2023r, gu2021densetnt, jiang2023motiondiffuser, li2022graph, fang2025neuralized, capellera2024footbots, capellera2025jointdiff, capellera2025unified, capellera2024transportmer, teoh2025coherent, capelleraTPAMI26}, and the other forecasts how their articulated body will move independently of where it is~\cite{hosseininejad2025motionmap, bouazizi2022motionmixer, choudhury2023tempo, xu2023auxiliary, martinez2021pose, sun2024comusion, medjaouri2022hr, graber2020dynamic, chen2023humanmac, barquero2023belfusion, saadatnejad2023generic, curreli2025nonisotropic}. However, in team sports, both quantities matter simultaneously and are far from independent: a player's path across the pitch and the way their limbs move at any instant are two views of the same underlying action, and practitioners care about both, trajectories inform tactical analysis, while pose informs technique and injury risk. Some works have begun to forecast pose and trajectory jointly~\cite{adeli2021tripod, jeong2024multi, agrawal2026simplihumon}, but unfortunately, they target settings with at most a couple of interacting people, where the social axis of motion is largely absent. Soccer offers a setting where this assumption breaks down completely: with twenty-plus players continuously reacting to teammates, opponents, and the state of play.

This work studies joint trajectory and pose forecasting for interacting groups of players found in real soccer matches.
As depicted in Fig.~\ref{fig:overview}, we adopt a socio-temporal transformer encoder, following the coarse-to-fine factorized attention design proposed for unimodal trajectory forecasting~\cite{capellera2024transportmer}, and apply it to a fused token stream that combines both modalities at the input level, letting a single encoder attend jointly over time and agents at once rather than predicting one and conditioning the other or processing the two streams through separate branches before merging. We further contribute two loss-level improvements, validated with targeted ablations: a per-joint rebalancing scheme with target joints selected via empirical per-joint error measurement, and a frequency-domain trajectory smoothing term that indirectly stabilizes pose forecasts as well.

Another aspect of this work concerns how we evaluate the results against existing methods. The most suitable datasets for this task, WorldPose~\cite{jiang2024worldpose} and 3DPW~\cite{vonMarcard20183dpw}, must be cut into short, fixed-length sequences before training and evaluation. We find that the specific protocol used by the most recent methods with public implementations tackling this same joint task, T2P~\cite{jeong2024multi} and SimpliHumon~\cite{agrawal2026simplihumon}, extract sequences with a stride of a single frame, introducing avoidable biases: near-identical sequences are counted many times over, and longer source videos receive disproportionate weight in the reported average simply by virtue of their length. When re-evaluating these two methods for direct comparison, we instead construct sequences with a non-overlapping protocol, and additionally adopt a deterministic single-future evaluation that does not require oracle access to the ground truth at inference time, unlike the best-of-many selection used in their original papers.

We evaluate our approach on WorldPose~\cite{jiang2024worldpose}, real broadcast footage from the 2022 FIFA World Cup featuring large groups of simultaneously interacting players, and on 3DPW~\cite{vonMarcard20183dpw}, a general-purpose benchmark in which each sequence contains at most two people. WorldPose~\cite{jiang2024worldpose} is the setting our socio-temporal design is most directly aimed at, and our model substantially improves over competing approaches across the majority of metrics. Notably, despite 3DPW~\cite{vonMarcard20183dpw} offering little of the social context our architecture is built to exploit, our model also matches or exceeds both baselines on this benchmark, indicating that the gains reported in this work do not depend on a densely populated scene and that the underlying joint formulation generalizes well beyond the team-sport setting that motivated it.

The main contributions of this work are as follows:
\begin{itemize}
    \item A socio-temporal encoder that fuses trajectory and 3D pose at the input level and processes them jointly through shared factorized attention, at the scale of a full soccer scene, with a controlled comparison of fusion topologies confirming that input-level fusion outperforms sequential and late alternatives under a matched backbone.
    \item A per-joint loss rebalancing scheme that directs optimization toward the most informative body joints, with target joints selected based on directly measured per-joint error under uniform weighting rather than by assumption.
    \item A frequency-based trajectory smoothing term that suppresses unrealistic high oscillations in the predicted path, improving trajectory accuracy and, as an additional benefit, the stability of the accompanying pose forecasts.
    \item State-of-the-art on WorldPose~\cite{jiang2024worldpose} and best among compared methods on 3DPW~\cite{vonMarcard20183dpw}, despite its markedly lower agent density, demonstrating effectiveness both in the large-group setting it targets and beyond.
\end{itemize}

\section{Related Work}
\label{sec:related}
Forecasting how people move and how their bodies articulate over time has been studied from several separate angles: predicting the global trajectory of one or several agents, predicting the temporal evolution of an articulated skeleton, and more recently predicting both jointly. We review each direction below, paying particular attention to which settings, datasets, and forecasting horizons each line of work targets, since these choices determine which methods constitute a meaningful point of comparison for the soccer setting addressed in this paper.

\noindent\textbf{Trajectory Forecasting} has been extensively studied in different types of agents moving through shared spaces~\cite{aydemir2023adapt, choi2023r, gu2021densetnt, jiang2023motiondiffuser, li2022graph}, with graph- and attention-based architectures now the dominant paradigm for modeling interactions between agents sharing a scene~\cite{yao2024trajclip, bae2024singulartrajectory, rowe2023fjmp, peral2026multi}. Within sports specifically, trajectory forecasting has been explored for American football~\cite{capellera2025jointdiff, teoh2025coherent}, basketball~\cite{fang2025neuralized, capellera2025jointdiff, capellera2025unified, teoh2025coherent}, and soccer~\cite{capellera2025unified, capellera2024transportmer, gutiACCV26}, taking advantage of the structured and rule-governed nature of team play to constrain the space of plausible futures. However, none of these methods forecast body pose alongside trajectory as we do in this work.

\noindent\textbf{Human Pose Forecasting.}
A separate body of work forecasts the future articulation of a skeleton without reasoning about its position in the world. Early approaches relied on recurrent architectures~\cite{hosseininejad2025motionmap, bouazizi2022motionmixer, choudhury2023tempo, xu2023auxiliary}, later replaced by feed-forward graph- and transformer-based models that predict an entire future sequence directly and avoid autoregressive error accumulation~\cite{martinez2021pose, sun2024comusion, medjaouri2022hr, graber2020dynamic}, as well as diffusion-based models that generate multiple plausible futures~\cite{chen2023humanmac, barquero2023belfusion, saadatnejad2023generic, curreli2025nonisotropic}. A parallel line of work operates in the frequency domain via the discrete cosine transform, decomposing pose sequences into a small set of low-frequency components that capture smooth, periodic motion patterns particularly well~\cite{rahman2023best, gao2023decompose, peng2023trajectory, wang2021multi}.

Most of these methods, including representative recent examples like MotionMixer~\cite{bouazizi2022motionmixer} and BeLFusion~\cite{barquero2023belfusion}, are formulated for a single person and report results on benchmarks like Human3.6M~\cite{h36m_pami} or AMASS~\cite{naureen2019amass}, where no inter-person interaction is modeled, typically over forecasting horizons of one second or less. Extensions to multi-person settings instead design attention mechanisms that operate jointly over agents and joints~\cite{tanke2023social, vendrow2022somoformer}, or combine local per-person encoders with a global scene representation~\cite{wang2021multi}.
These models, however, are trained on substantially larger pose corpora than either of our target benchmarks provides: \cite{wang2021multi}, for instance, trains exclusively on CMU-Mocap~\cite{cmu2003} before testing zero-shot on 3DPW~\cite{vonMarcard20183dpw} and MuPoTS-3D~\cite{mehta2018single}. More importantly, none of these multi-person methods output a global trajectory, so predictions remain root-relative and the agent’s position in the scene is never recovered.
EqMotion~\cite{xu2023eqmotion} applies the same multi-agent reasoning machinery to both tasks, treating joints as interacting agents to forecast pose and pedestrians as interacting agents to forecast trajectory. The issue is that it evaluates the two as separate scenarios rather than predicting, for the same set of people, how their bodies articulate and where they move at once, which is the setting we address here.

\noindent\textbf{Joint Pose and Trajectory Forecasting.}
A smaller body of work forecasts pose and trajectory together, motivated by the observation that a person's global movement and the articulation of their limbs are tightly coupled, a runner's stride pattern is inseparable from their forward motion, and a model that treats the two as independent discards information relevant to both.
On that same premise, a complementary line of research treats body pose as an auxiliary input that sharpens future trajectory forecasting~\cite{saadatnejad2023social, gao2025omnitraj, gao2024multi, gao2025social, gutierrez2026once, peral2026synergy}.
Social-Transmotion~\cite{saadatnejad2023social} fuses pose-keypoint tokens with trajectory tokens at the input of a shared social transformer, using masking to remain robust to unavailable modalities. Nonetheless, the method forecasts only trajectories. Multi-Transmotion~\cite{gao2024multi} extends this into a large multimodal pre-trained model that also emits 3D pose keypoints. However, it forecasts the future motion of a single ego agent while treating the remaining agents purely as social context, being trajectory and pose capabilities pre-trained on seven merged datasets and evaluated on disjoint benchmarks, never providing the joint evaluation on a common testbench.

Early architectures shared representations across the two tasks through multi-branch designs~\cite{adeli2021tripod, cao2020long, zaier2023dual}, while later work explored sequential pipelines in which a trajectory is forecast first and used to condition pose generation~\cite{parsaeifard2021learning, jeong2024multi}. More recent sequence-based methods operate directly on structured pose inputs, separating and later fusing local and global representations~\cite{zheng2025efficient} or introducing learnable query tokens to represent candidate future motion states~\cite{agrawal2026simplihumon}, with both lines of work focused primarily on improving temporal reasoning within pose-centric representations.

Among methods that jointly forecast pose and trajectory for multiple interacting people, TRiPOD~\cite{adeli2021tripod} is closely related to our setting, modeling human-human and human-object interactions through graph attention and evaluating on PoseTrack~\cite{andriluka2018posetrack} and 3DPW~\cite{vonMarcard20183dpw} over a short forecast horizon of 900ms. Unfortunately, at the time of writing, no official implementation has been released, which precludes a direct experimental comparison. We instead compare against T2P~\cite{jeong2024multi} and SimpliHumon~\cite{agrawal2026simplihumon}, which, unlike the methods discussed above, are trained directly on our target benchmarks without relying on larger external pretraining datasets, support forecasting horizons beyond one second, and tackle the same pair of tasks, pose and trajectory forecasting, that we address. These properties make them the most directly comparable existing methods, and we adopt them as our primary baselines. However, we identify a methodological issue in how both construct sequences from 3DPW~\cite{vonMarcard20183dpw}: their official protocols use a stride of a single frame across all splits, a choice we revisit and modify due to it introducing biases, and we propose instead a non-overlapping protocol. Our approach departs from the methods discussed in this section by fusing trajectory and pose at the input level and processing them through a shared socio-temporal encoder, without imposing a sequential or branch-separated coupling between the two modalities.

\section{Methodology}
\label{sec:method}

\subsection{Problem Statement}
\label{ssec:problem}

Given a set of $N \in \mathbb{N}$ agents observed over $T$ frames, each described by motion and body pose features, our goal is to use the first $t < T$ observations to predict the remaining frames, i.e., from $t+1$ to $T$. For each observed time step, the model receives two input streams: player trajectories $\boldsymbol{x}_t \in \mathbb{R}^{N \times 2}$, given as 2D positions in global pitch coordinates~\cite{gutierrez4998149pnlcalib}, and 3D body poses $\boldsymbol{p}_t \in \mathbb{R}^{N \times J \times 3}$, where $J$ denotes the number of joints. Poses are represented as joint-wise offsets relative to a root joint fixed at the origin, which isolates articulated motion from global position; the trajectory stream separately captures this positional information, making the two modalities complementary by construction. The model operates under a strictly causal setting, relying exclusively on past observations to forecast future ones, with no access to future context at inference time. Predicted trajectories and poses are expressed in the same coordinate systems as their respective inputs, ensuring consistency between observation and prediction spaces.

\subsection{Multi-Modal Input Encoding}
\label{ssec:input}

Soccer scenes contain a variable number of visible players due to occlusions and incomplete tracking. Instead of propagating attention masks throughout the encoder, we pad missing agent slots with a fixed out-of-range sentinel value (e.g., $-2000$), which cannot overlap with valid normalized trajectory or pose values. This architecture-agnostic strategy allows the model to process up to $N$ agents without conditional logic. In practice, attention weights assigned to padded slots converge close to zero, indicating that the model implicitly learns to ignore absent agents.
As shown in Fig.~\ref{fig:overview}, trajectory and pose inputs are first projected independently into a shared latent space of dimension $d$, producing two $t \times N \times d$ representations. These are concatenated, yielding a $t \times N \times 2d$ tensor, and fused through a small MLP that compresses the representation back to $d$ dimensions. This early fusion enables cross-modal interactions from the first encoder layers while maintaining a consistent embedding size, jointly encoding each agent's position and body configuration at every timestep.

\subsection{Socio-Temporal Encoder}
\label{ssec:encoder}

To process the fused multimodal representation, we adopt the coarse-to-fine socio-temporal encoder introduced in~\cite{capellera2024transportmer} for trajectory forecasting, which we extend here to operate over our joint trajectory-pose tokens. The encoder consists of two sequentially stacked stages, a coarse and a fine one, that share the same internal structure but act on progressively refined representations: the first builds a global picture of the scene, while the second resolves finer-grained dependencies on top of it (see Fig.~\ref{fig:overview}). Before the first stage, a temporal positional encoding is added to the token sequence to preserve frame ordering, while no positional encoding is applied along the agent dimension, which enforces permutation equivariance with respect to player ordering and lets the model generalize across different team compositions. In our work each stage is composed of three Set Attention Blocks (SAB)~\cite{capellera2024transportmer,lee2019set} arranged as two temporal blocks, $SAB_T$, followed by one social block, $SAB_S$. Each $SAB_T$ attends independently over the time dimension for a given agent, aggregating its individual motion history, while the subsequent $SAB_S$ attends across all agents within a single frame, capturing inter-player interactions and collective team behavior. We adopt this factorized design as an efficient and already validated way of scaling attention to the large number of simultaneously interacting agents present in a soccer scene, and our contribution lies not in this backbone itself but in fusing trajectory and pose into a shared token stream that this encoder then processes jointly. 

\subsection{Prediction Heads}
\label{ssec:heads}

The output of the fine encoder is passed to two task-specific prediction heads, each implemented as an independent MLP operating on the shared spatio-temporal representation. The trajectory head maps the encoder output to a tensor of shape $T \times N \times 2$, the predicted 2D pitch-space position of each agent at each timestep, while the pose head maps the same representation to a tensor of shape $T \times N \times 3J$, the predicted 3D root-relative joint coordinates for each player. In both cases, only the forecast frames, from $t+1$ to $T$, are retained as the final output, discarding the reconstructions produced for the observed frames at inference time.

\subsection{Loss}
\label{ssec:losses}

The model is trained end-to-end using a composite loss that jointly supervises trajectory and pose forecasting, together with a smoothness term applied to the predicted trajectory. The forecasting terms are computed over every frame, including the observed ones, training the model in a sequence-to-sequence manner. Supervising the reconstruction of observed frames alongside future prediction regularizes the temporal dynamics learned by the encoder and encourages consistent representations across the observation-prediction boundary. The total loss is written as:
\begin{equation}
    \mathcal{L}_\text{total} = \mathcal{L}_\text{traj} + \lambda_\text{pose} \mathcal{L}_\text{pose} + \lambda_\text{smooth} \mathcal{L}_\text{smooth},
    \label{eq:loss}
\end{equation}
\noindent where $\lambda_\text{pose}$ and $\lambda_\text{smooth}$ control the relative contribution of each term. We set $\lambda_\text{pose}=\lambda_\text{smooth}=10$ throughout our experiments. Each term is detailed below.

\noindent \textbf{Trajectory Loss.}
The trajectory term penalizes the mean Euclidean distance between predicted and ground-truth pitch positions across all agents and frames:
\begin{equation}
  \mathcal{L}_\text{traj} = \frac{1}{NT} \sum_{n=1}^N \sum_{t=1}^T \left\| \boldsymbol{\hat{x}}_{n,t} - \boldsymbol{x}_{n,t} \right\|_2 ,
    \label{eq:losstraj}
\end{equation}
\noindent where $\boldsymbol{\hat{x}}_{n,t}$ and $\boldsymbol{x}_{n,t}$ are the predicted and ground-truth positions of player $n$ at frame $t$, respectively.

\noindent \textbf{Pose Loss.}
The pose term supervises the predicted root-relative joint coordinates across all agents, frames, and joints. For each $j$-th joint of player $n$ at frame $t$, we first compute an absolute error as $\epsilon_{n,t,j} = | \hat{p}^x_{n,t,j} - p^x_{n,t,j}| + |  \hat{p}^y_{n,t,j} - p^y_{n,t,j}| + |  \hat{p}^z_{n,t,j} - p^z_{n,t,j}| $, where $\hat{p}_{n,t,j}$ and $p_{n,t,j}$ denote the predicted and ground-truth 3D coordinates, respectively. We then take the $l_2$ norm over the full set of joints, which acts as a regularizer that heavily penalizes large localized errors, such as those arising from rapid limb movements, that would otherwise be concealed by the accurate prediction of more stable joints. A small subset of distal joints, namely the elbows, wrists, ankles, and feet, exhibit a markedly higher error than the rest of the skeleton under uniform loss weighting, since their displacement in root-relative coordinates is substantially larger and harder to predict than that of proximal joints such as the hips or torso. Guided directly by that, we introduce a per-joint weight vector $\boldsymbol{w} \in \mathbb{R}^J$ that up-weights precisely this set of high-error distal joints, preventing them from being dominated in the loss by the larger number of easier, low-error proximal joints, and explicitly encouraging the model to learn their fine-grained articulation. The pose loss is then written as:
\begin{equation}
  \mathcal{L}_\text{pose} = \frac{1}{NT} \sum_{n=1}^N \sum_{t=1}^T \left\| \boldsymbol{w} \odot \boldsymbol{\epsilon}_{n,t} \right\|_2 ,
    \label{eq:losspose}
\end{equation}
\noindent where $\boldsymbol{\epsilon}_{n,t} = (\epsilon_{n,t,1}, \ldots, \epsilon_{n,t,J}) \in \mathbb{R}^J$ collects the per-joint errors, and $\odot$ denotes an element-wise product. This design discourages degenerate solutions in which the model collapses toward static average poses, instead incentivizing the learning of the high-intensity, fine-grained motion patterns essential to realistic soccer dynamics. The relative strength of this rebalancing is controlled by a single scalar applied to the weights of the selected joints, found empirically.

\noindent \textbf{Trajectory Smoothing Loss.}
Agent positions predicted independently at each frame can exhibit unrealistic high-frequency oscillations, or zigzagging, between consecutive timesteps, even when their average displacement error is low. To discourage this behavior, we introduce a smoothness term that penalizes high-frequency content in the predicted trajectory directly in the frequency domain, so that any systematic oscillation occurring anywhere within the forecasting horizon is captured, rather than only oscillations between immediately adjacent frames. For each agent, we take the real-valued fast Fourier transform of its predicted trajectory over the $T_{pred}$ forecast frames independently along each coordinate axis, obtaining a complex spectrum with $F = \lfloor T_{pred}/2 \rfloor + 1$ frequency bins. We then compute the power spectrum as the squared magnitude of each frequency coefficient and weight it by the square of its normalized frequency, so that the static component receives zero weight and the highest representable Nyquist frequency receives the maximum weight of one as:
\begin{equation}
  \mathcal{L}_\text{smooth} = \frac{1}{N F C} \sum_{n=1}^{N} \sum_{k=0}^{F-1} \sum_{c=1}^{C} f_k^2 \left| \hat{q}_{n,k,c} \right|^2 ,
  \label{eq:losssmooth}
\end{equation}
where $C=2$ is the number of spatial coordinates, $\hat{q}_{n,k,c}$ is the $k$-th Fourier coefficient of the $c$-th coordinate of agent $n$'s predicted trajectory, and $f_k = k / (F-1) \in [0,1]$ is the normalized frequency of bin $k$. By penalizing high-frequency energy, this term suppresses zigzagging and other trajectory artifacts while largely preserving physically plausible motion. Moreover, it consistently improves pose forecasting by providing a steadier trajectory context.

\section{Experimental Results}
\label{sec:experiments}

We now present both quantitative and qualitative results in terms of human pose and trajectory prediction. First, we describe the datasets, evaluation protocol, metrics and implementation details, followed by a series of ablation studies assessing the contribution of each block. Finally,
we compare our method with respect to two recent methods.

\subsection{Datasets, Metrics and Implementation details}
\label{ssec:datasets}
We evaluate our approach on two datasets: WorldPose~\cite{jiang2024worldpose} and 3DPW~\cite{vonMarcard20183dpw}. On one side, WorldPose~\cite{jiang2024worldpose} provides precise 3D pose annotations from broadcast footage of the 2022 FIFA World Cup, spanning roughly 50 minutes of play across multiple matches and venues, with every player visible on the pitch annotated simultaneously. We segment the recordings into disjoint 3-second windows for training and evaluation, without temporal overlap between extracted sequences. Unlike the original benchmark protocol, where train, validation, and test sequences can be drawn from the same match, we enforce a match-level split so that no evaluation sequence shares a video source with any training sequence, removing incidental cues that would otherwise leak across splits and give an overly optimistic picture of generalization to unseen games. We use the Morocco-Portugal as validation, Netherlands-Argentina as test, and the other 6 games for training. On the other hand, 3DPW~\cite{vonMarcard20183dpw} is captured with a moving camera across diverse indoor and outdoor scenarios and typically involves at most two people per clip. The recordings are provided at 30 fps, so, to match the 3-second duration used for WorldPose~\cite{jiang2024worldpose}, we extract sequences of 90 consecutive frames, and we keep the per-video split of the original benchmark.

Both datasets have served prior work on many pose-related tasks, but for the specific problem of forecasting future trajectory and pose, the protocol used by methods tackling it~\cite{jeong2024multi, agrawal2026simplihumon} slides the extraction window forward one frame at a time. Under this unit stride, adjacent windows overlap almost completely, so each stretch of motion is scored many times, appearing both as a forecast target and as observed context for numerous neighboring windows. Longer recordings also yield proportionally more windows than shorter ones. Together, these effects let a small number of recurring motions and the longest recordings dominate the reported average rather than the metric reflecting an even sample over the dataset's diversity.
For both datasets we extract windows with a non-overlapping stride, partitioning each split's frames into disjoint, equal-length sequences so that every frame contributes to exactly one, and discarding any trailing remainder shorter than a full window. Each sequence provides one second of observed context and is evaluated on the following two seconds, the configuration used in all experiments.

\noindent\textbf{Metrics and implementation details.} We report five complementary metrics, two per modality and a shared one, all in mm. For trajectories, Average Displacement Error (ADE) is the mean Euclidean distance between predicted and ground-truth 2D root positions over all forecast frames, while Final Displacement Error (FDE) measures the same error only at the final frame, capturing end-of-horizon drift.
For pose, Aligned mean per joint Pose Error (APE) is the mean Euclidean joint error after expressing predictions and ground truth in the root-joint coordinate system, averaged over joints and frames, thus isolating articulation from global positioning. Procrustes-aligned mean per joint Pose Error (PPE) is computed analogously after full Procrustes alignment, including rotation and scaling. Finally, to score the joint task directly, we report the mean per-joint error in world coordinates (JPE). We place each predicted root-relative pose at the predicted pitch position, assuming the vertical coordinate as the trajectory head predicts only 2D positions, and measure the mean Euclidean distance to the ground-truth joints in the global frame, averaged over all joints and forecast frames. Unlike ADE/FDE (which see only the root) and APE/PPE (which are root-relative or Procrustes-aligned), JPE is invariant to neither modality and so reflects the coupled accuracy of trajectory and pose together. Nevertheless, being an absolute world-frame distance, it is numerically dominated by the global-position term.
Our model is trained end-to-end for 100 epochs on a single NVIDIA A40 GPU, using the AdamW optimizer with a learning rate of $10^{-3}$ and $\epsilon = 10^{-4}$, a batch size of 32, embedding dimension of $d=128$, and gradient norm clipping at 5.

\subsection{Ablation Studies}
\label{ssec:ablations}

\noindent\textbf{Effect of Cross-Modal Inputs.}
We first isolate how much each input modality contributes to the other's forecasting task, by training single-task variants of our model that predict only trajectories or only poses, and feeding them either their own matching modality alone or both modalities jointly. Table~\ref{tab:pose_input_ablation} reports the pose/trajectory-only setting, respectively. When predicting poses alone (left part in the table), supplying trajectory as an additional input improves both pose metrics, even though no trajectory is being forecast in this configuration. This confirms that knowing where a player is heading on the pitch is informative for anticipating how their body will be configured. However, the effect is considerably larger when predicting trajectories alone: adding pose as an input substantially reduces both ADE and FDE, indicating that the articulation of a player's limbs carries strong cues about their imminent movement, such as the pattern of an accelerating run, which a model can exploit to forecast position more accurately. Beyond the benefit of richer inputs, we also find that training the two tasks jointly, rather than training single-task variants, as in this ablation, gives a further improvement over either single-task model. This is especially pronounced for pose forecasting: once the model is also tasked with predicting future trajectories, the representations it must learn to do so reflect a coherent notion of where the player is going, and this improved trajectory context in turn sharpens pose forecasts beyond what is achieved by simply observing trajectory as a static input. The joint formulation used throughout the rest of this paper therefore benefits from both the cross-modal input signal studied here and the additional supervision that comes from forecasting both modalities at once.

\begin{table}[t!]
\centering
\begin{minipage}{0.48\linewidth}
\centering
\resizebox{\linewidth}{!}{
\begin{tabular}{c|cc}
\toprule
\textbf{Input Modalities} & \textbf{APE} $\downarrow$ & \textbf{PPE} $\downarrow$ \\
\hline
Pose only & 123.0 & 90.7 \\
Pose + Trajectory & \textbf{121.8} & \textbf{88.9} \\
\bottomrule
\end{tabular}
}
\end{minipage}
\hfill
\begin{minipage}{0.48\linewidth}
\centering
\resizebox{\linewidth}{!}{
\begin{tabular}{c|cc}
\toprule
\textbf{Input Modalities} & \textbf{ADE} $\downarrow$ & \textbf{FDE} $\downarrow$ \\
\hline
Trajectory only & 901.7 & 1901.5 \\
Trajectory + Pose & \textbf{704.1} & \textbf{1681.4} \\
\bottomrule
\end{tabular}
}
\end{minipage}
\vspace{-0.2cm}
\caption{\textbf{Effect of trajectory and pose as an auxiliary input on pose/trajectory-only forecasting}, evaluated on WorldPose~\cite{jiang2024worldpose}. The model is trained to predict poses (left) / trajectories (right) exclusively; only the input modalities are varied.}
\label{tab:pose_input_ablation}
\vspace{-0.2cm}
\end{table}
\begin{table}[t!]
\centering
\resizebox{8.2cm}{!} {
\begin{tabular}{cc|ccccc}
\toprule
\textbf{Fusion} & \textbf{Params} & \textbf{ADE} $\downarrow$ & \textbf{FDE} $\downarrow$ & \textbf{JPE} $\downarrow$ & \textbf{APE} $\downarrow$ & \textbf{PPE} $\downarrow$ \\
\hline
Sequential      & 2.58M & 1074.6 & 2186.7 & 1131.9 & 116.3 & 85.3 \\
Cross Att & 1.45M & 951.4  & 1962.8 & 1012.7 & 115.1 & 84.1 \\
Late Halved     & 0.74M & 850.9  & 1875.2 & 907.3  & 115.2 & 84.4 \\
Late            & 2.58M & 810.7  & 1804.9 & 863.6  & 114.8 & 84.2 \\
Early Sum       & 1.31M & 681.2  & 1663.4 & 736.8  & 113.0 & 83.9 \\
Early Concat    & 1.32M & \textbf{670.8}  & \textbf{1642.0} & \textbf{723.5}  & \textbf{112.6} & \textbf{83.7} \\
\bottomrule
\end{tabular}}
\vspace{-0.2cm}
\caption{\textbf{Fusion topology} on WorldPose. All lines share backbone, losses, protocol and training recipe, only the fusion point changes. Rebalancing/smoothing are disabled to isolate fusion.}
\label{tab:fusion}
\end{table}


\noindent\textbf{Fusion Topology.} Before studying our loss terms, we first establish where the two modalities should be combined, holding the backbone, losses, protocol and training recipe fixed and varying only the fusion point. Table~\ref{tab:fusion} compares five alternatives: a sequential design that forecasts trajectory first and conditions pose on it; a cross-attention block in which pose and trajectory tokens attend to one another before the encoders; two-branch late fusion, in which each modality is processed by its own encoders and merged only before the output heads, at full width and at halved width to contrast the parameter budget; input-level fusion of the two modalities into a single per-agent, per-frame token, either by element-wise sum or by concatenation followed by an MLP (ours). Input-level fusion is clearly strongest, and the separation between designs is concentrated in the trajectory-coupled metrics while the root-relative pose metrics stay nearly flat.
The sequential prior is weakest by a wide margin, consistent with the cross-modal analysis in Table~\ref{tab:pose_input_ablation} where pose$\rightarrow$trajectory is the stronger direction, confirming that forecasting trajectories without access to pose modality is unfavorable. 
Finally, halved late fusion trails full late fusion, while early concatenation surpasses it with roughly half the parameters, pointing to a topological rather than capacity advantage. Concatenation slightly outperforms summation, so we use it throughout.

\noindent\textbf{Joint Balancing.}
We next study the effect of the joint rebalancing weight (see Section~\ref{ssec:losses}), which consistently up-weights the same set of high-motion joints previously selected based on their measured error, varying only the strength of this rebalancing. Table~\ref{tab:ablations}-top reports trajectory and pose metrics as the strength increases, with trajectory smoothing disabled throughout so that its effect is isolated from the smoothing term studied next. By steering the loss toward the most dynamic joints, this rebalancing helps the model escape the local minimum of predicting a near-static skeleton, a degenerate solution into which a uniformly weighted loss can otherwise settle since low-motion joints dominate the average error. Increasing the rebalancing factor steadily improves pose metrics, confirming that directing more of the loss toward the most dynamic joints sharpens articulation. This comes at a cost to trajectory accuracy, which degrades as the factor grows, since the trajectory term receives comparatively less effective weight as optimization is increasingly steered toward pose.
We select a factor of $5$ for the rest of our experiments, as it lies close to the point where further pose gains begin to come at a disproportionate cost to trajectory accuracy.
JPE, which aggregates both terms, happens to be minimized at this same factor (with and without smoothing), rising for larger factors as the growing trajectory error dominates the sum. We report this only as a consistency check rather than as the basis for the choice, since JPE is dominated by the global position term and is correspondingly insensitive to the local articulation that the rebalancing is meant to improve.

\begin{table}[t!]
\centering
\resizebox{8.2cm}{!} {
\begin{tabular}{cc|ccccc}
\toprule
\textbf{Joints} & \textbf{Smoothing} & \textbf{ADE} $\downarrow$ & \textbf{FDE} $\downarrow$ & \textbf{JPE} $\downarrow$ & \textbf{APE} $\downarrow$ & \textbf{PPE} $\downarrow$ \\
\hline
0  & \XSolidBrush  & 670.8 & 1642.0 & 723.5 & 112.6 & 83.7 \\
2.5 & \XSolidBrush & 676.3 & 1651.1 & 720.2 & 105.9 & 81.4 \\
5  & \XSolidBrush  & 683.3 & 1667.2 & 719.8 & 101.3 & 79.1 \\
7.5 & \XSolidBrush & 714.6 & 1694.0 & 745.6 & 99.8  & 78.7 \\
10  & \XSolidBrush & 848.5 & 1832.5 & 874.7 & 98.7  & 78.5 \\
\hline
0 & \Checkmark   & 611.7 & 1528.1 & 659.1 & 106.7 & 82.6 \\
5 & \Checkmark   & 623.0 & 1549.1 & 647.3 & 95.2  & 77.7 \\
\bottomrule
\end{tabular}
}
\vspace{-0.2cm}
\caption{\textbf{Ablation studies} evaluated on WorldPose~\cite{jiang2024worldpose}. \textbf{Joint rebalancing factor} in the top rows show that larger factors steadily improve pose accuracy at the expense of trajectory accuracy. \textbf{Trajectory smoothing} enabled for the last two rows improves trajectory accuracy directly, and pose accuracy indirectly, by removing unrealistic high-frequency jitter in the predictions}
\vspace{-0.2cm}
\label{tab:ablations}
\end{table}

\begin{figure*}[t!]
    \centering
    \includegraphics[width=0.9\textwidth]{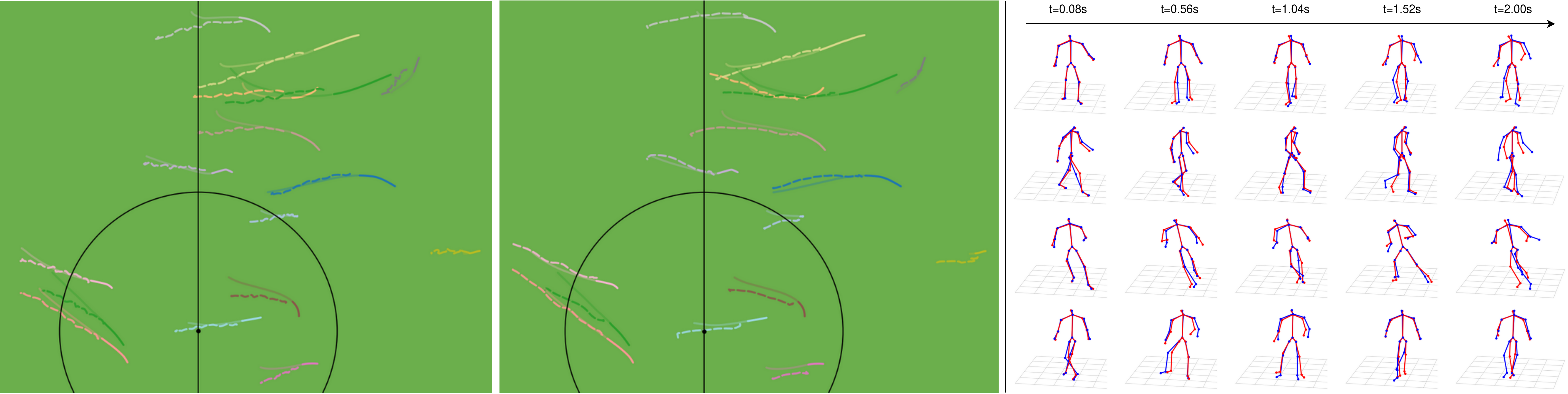}
    \vspace{-0.3cm}
    \caption{\textbf{Effect of trajectory smoothing on predicted player paths.} \textbf{Left and central part:} In each crop, the solid opaque line shows the observed prior trajectory, the solid semi-transparent line shows the ground-truth future trajectory, and the dashed opaque line shows the predicted path. Left and central representations are without and with the smoothing term enabled, respectively.
    \textbf{Right part:} Qualitative pose forecasting on WorldPose~\cite{jiang2024worldpose} test set. Ground-truth poses are shown in blue and predicted poses in red. Each row shows a single player action at five temporally equidistant frames within the 2-second forecasting horizon. Best viewed with zooming.}
    \label{fig:smoothing_traj}
    \vspace{-0.3cm}
\end{figure*}

\noindent\textbf{Trajectory Smoothing.}
We then evaluate the frequency-domain smoothing term (see Section~\ref{ssec:losses}), which penalizes high-frequency components of the predicted trajectory to discourage unrealistic wiggling between consecutive frames. Last rows of Table~\ref{tab:ablations} show models trained with this term, to be compared with their analogs with the same joints balancing factor without smoothing in the top rows. Enabling smoothing improves trajectory accuracy in both cases, as expected given its direct effect on the trajectory loss. Notably, it also yields a consistent improvement in both pose metrics: by discouraging zigzagging trajectory predictions, the model produces a more physically coherent notion of where each player is heading, and this steadier trajectory context translates into more stable and realistic pose forecasts, reinforcing the cross-modal dependency already observed in Table~\ref{tab:pose_input_ablation}. Figure~\ref{fig:smoothing_traj} left and central crops illustrate this effect qualitatively. Without the smoothing term (left), predicted trajectories exhibit visible jitter from frame to frame, deviating back and forth around the underlying path rather than tracing it cleanly. With smoothing enabled (central), the same trajectories become clearly steadier. This added regularity is not merely cosmetic: by removing the spurious oscillation, the predicted path is also better at following the true curvature of the trajectory, since a smoother forecast can commit to a consistent direction of turning instead of having that signal masked by frame-to-frame noise.

\begin{table*}[t!]
\centering
\resizebox{16.2cm}{!} {
\begin{tabular}{c|c|ccccc|ccccc}
\toprule
\multirow{2}{*}{\textbf{Method}} & \multirow{2}{*}{\textbf{Futures}} & \multicolumn{5}{c|}{\textbf{WorldPose~\cite{jiang2024worldpose}}} & \multicolumn{5}{c}{\textbf{3DPW~\cite{vonMarcard20183dpw}}} \\
 & & \textbf{ADE} $\downarrow$ & \textbf{FDE} $\downarrow$ & \textbf{JPE} $\downarrow$ & \textbf{APE} $\downarrow$ & \textbf{PPE} $\downarrow$ & \textbf{ADE} $\downarrow$ & \textbf{FDE} $\downarrow$ & \textbf{JPE} $\downarrow$ & \textbf{APE} $\downarrow$ & \textbf{PPE} $\downarrow$ \\
\hline
StaticBaseline & -- & -- & -- & -- & 169.8 & 124.3 & -- & -- & -- & 108.5 & 79.5 \\
\hline
SimpliHuMoN~\cite{agrawal2026simplihumon} & 1 & 757.0 & 1920.3 & 803.4 & 157.3 & 109.2 & 613.0 & 1155.5 & 693.4 & 191.6 & 139.4 \\
SimpliHuMoN~\cite{agrawal2026simplihumon} & 6 & 763.0 & 1988.3 & 804.4 & 149.8 & 106.7 & 615.6 & 1161.5 & 702.2 & 202.2 & 174.8 \\
\hline
T2P~\cite{jeong2024multi} & 1 & 584.6 & 1668.0 & 631.6 & 142.2 & 94.4 & 636.8 & 1156.9 & 667.5 & 109.7 & 80.5 \\
T2P~\cite{jeong2024multi} & 6 & \textbf{564.1} & 1625.4 & \textbf{613.0} & 142.1 & 94.1 & 650.1 & 1165.5 & 682.4 & 109.1 & 80.2 \\
\hline
\textbf{Ours} & 1 & 623.0 & \textbf{1549.1} & 647.3 & \textbf{95.2} & \textbf{77.7} & \textbf{551.2} & \textbf{763.8} & \textbf{580.7} & \textbf{101.2} & \textbf{76.2} \\
\bottomrule
\end{tabular}}
\vspace{-0.2cm}
\caption{\textbf{Quantitative comparison against T2P~\cite{jeong2024multi}, SimpliHuMoN~\cite{agrawal2026simplihumon}, and a static baseline on WorldPose~\cite{jiang2024worldpose} and 3DPW~\cite{vonMarcard20183dpw}, with 1 second of observed context and a 2 second forecasting horizon.} \textit{Futures} denotes the number of sampled continuations used at inference for generative baselines, averaged when greater than one. Our model achieves competitive results on WorldPose~\cite{jiang2024worldpose} and, despite the limited social context available, the best results on 3DPW~\cite{vonMarcard20183dpw}.}
\label{tab:main_comparison}
\vspace{-0.4cm}
\end{table*}

\subsection{Quantitative Results}
\label{ssec:quantitative}
We compare against two recent multi-person pose and trajectory forecasting methods, T2P~\cite{jeong2024multi} and SimpliHuMoN~\cite{agrawal2026simplihumon}, alongside a simple non-learned baseline. 
Both methods report results on 3DPW~\cite{vonMarcard20183dpw} under the unit-stride sequence construction discussed above. To make a fair comparison, we instead re-run their official implementations under our non-overlapping protocol and consistent 1s/2s observation-forecast split, on both WorldPose~\cite{jiang2024worldpose} and 3DPW~\cite{vonMarcard20183dpw}. T2P~\cite{jeong2024multi} and SimpliHumon~\cite{agrawal2026simplihumon} can sample several plausible continuations per sequence: their original evaluation protocol retains the candidate closest to the ground truth, which requires access to the future itself and is not realizable at deployment time. Since our goal is a deterministic, directly usable forecasting model we report these baselines under two settings that do not require oracle selection: drawing a single future, and drawing multiple futures from which we select, without any access to the ground truth, the one closest to their average, that is, the most representative or central sample, and report its error against the ground truth. We sample six futures in this setting, matching the number of futures used in the original source of both methods. Both settings reflect a selection strategy that would actually be available at inference time, in contrast to the best-of-many oracle selection used in the original evaluation protocol of these methods.

We additionally report a non-parametric static baseline that holds the last observed pose fixed for every future frame, re-centering it at each timestep so that its root joint matches the ground-truth root trajectory at that frame, since this baseline performs no trajectory prediction of its own. This isolates how much of the apparent forecasting difficulty is solved trivially by assuming no further articulation, and gives a lower bound of difficulty against which learned methods should be judged. Our model used for comparison in this section incorporates the joint rebalancing factor and trajectory smoothing term selected in Section~\ref{ssec:ablations}. Table~\ref{tab:main_comparison} reports the comparison across both datasets.
On WorldPose~\cite{jiang2024worldpose}, our model remains competitive with T2P~\cite{jeong2024multi} on ADE and JPE while improving all other metrics, with a clear pose-metric margin over both benchmarks and the static reference. Unlike competing methods, whose trajectories increasingly drift from ground truth with horizon, ours remain comparatively stable, yielding a smaller ADE–FDE gap. JPE closely follows global-position error, and T2P~\cite{jeong2024multi} retains a slight edge, mirroring its narrow ADE advantage, while its larger pose errors are masked by JPE's position-dominated scale. We therefore interpret JPE alongside the marginal metrics, which disentangle global localization from articulation. On 3DPW~\cite{vonMarcard20183dpw}, despite the scene containing at most two agents and therefore offering little social context, our model still achieves the best result on every metric. We also note that the static baseline lands remarkably close to the best solutions on this dataset, in contrast with the larger gap between them observed on WorldPose~\cite{jiang2024worldpose}. This is consistent with the nature of the two datasets: WorldPose~\cite{jiang2024worldpose} sequences are drawn from high-intensity match play where players are rarely stationary for long, so freezing the last pose is a poor predictor of the future, whereas 3DPW~\cite{vonMarcard20183dpw} contains a broader mix of everyday, lower-intensity, and often near-static activities, where holding the last observed pose is frequently a reasonable approximation of what comes next, leaving comparatively little room for a learned model to improve upon it.

\subsection{Qualitative Evaluation}
\label{ssec:qualitative}

We further inspect our pose forecasts in Fig.~\ref{fig:smoothing_traj}-right, showing four sequences at five equidistant frames across the 2-second horizon (0.08s, 0.56s, 1.04s, 1.52s, and 2.00s after the last observed frame). The first two sequences depict players walking at a relaxed pace, completing roughly two steps over the displayed window. In both cases, predictions stay close to ground truth, with residual error concentrated in body orientation at the furthest frames rather than limb configuration. The last two sequences correspond to substantially harder cases. In the third row, a player begins a sprint abruptly at the observation boundary, a sudden change in pace with no corresponding cue in the observed history, yet the forecast remains visually consistent and plausible. The fourth row shows a player walking sideways with crossing legs, an unusual gait pattern that is likely underrepresented in the training data and that we consider a feasible source of model limitations. Even there the resulting prediction remains reasonable, suggesting that the model degrades gracefully even on motion patterns it has seen comparatively rarely.

\section{Conclusion}
\label{sec:conclusion}

We presented a unified approach for joint trajectory and 3D pose forecasting in real soccer matches, fusing both modalities at the input through a shared socio-temporal encoder that learns their interaction directly. Per-joint loss reweighting and frequency-domain trajectory smoothing further improve pose and trajectory accuracy respectively, the latter also stabilizing pose forecasts through steadier positional context. Evaluated fairly with a non-overlapping, oracle-free protocol, our model achieves state-of-the-art results on WorldPose~\cite{jiang2024worldpose} and outperforms competing methods on 3DPW~\cite{vonMarcard20183dpw} despite its lower social density, with qualitative results confirming plausible forecasts under abrupt pace changes and graceful degradation on rare motions.

\noindent \textbf{Acknowledgment.} This work has been partially supported by the project GRAVATAR PID2023-151184OB-I00 funded by MCIU/AEI/10.13039/501100011033 and by ERDF, UE and by the Government of Catalonia under 2020 DI 00106.

{
    \small
    \bibliographystyle{ieeenat_fullname}
    \bibliography{main}

@String(CVPR= {IEEE Conf. Comput. Vis. Pattern Recog.})

@String(ICCV= {Int. Conf. Comput. Vis.})

@String(ECCV= {Eur. Conf. Comput. Vis.})

@String(ICIP = {IEEE Int. Conf. Image Process.})

@String(ACCV  = {ACCV})

@String(ICLR = {Int. Conf. Learn. Represent.})

@String(CVPR  = {CVPR})

@String(ICCV  = {ICCV})

@String(ECCV  = {ECCV})

@String(ICIP  = {ICIP})

@String(ICLR  = {ICLR})

@inproceedings{vonMarcard20183dpw,
title = {Recovering Accurate {3D} Human Pose in The Wild Using IMUs and a Moving Camera},
author = {von Marcard, Timo and Henschel, Roberto and Black, Michael and Rosenhahn, Bodo and Pons-Moll, Gerard},
booktitle = {ECCV},
year = {2018}
}

@article{agrawal2026simplihumon,
  title={SimpliHuMoN: Simplifying Human Motion Prediction},
  author={Agrawal, Aadya and Schwing, Alexander},
  journal={arXiv preprint arXiv:2603.04399},
  year={2026}
}

@inproceedings{mehta2018single,
  title={Single-shot multi-person 3d pose estimation from monocular rgb},
  author={Mehta, Dushyant and Sotnychenko, Oleksandr and Mueller, Franziska and Xu, Weipeng and Sridhar, Srinath and Pons-Moll, Gerard and Theobalt, Christian},
  booktitle={3DV},
  pages={120--130},
  year={2018}
}

@conference{naureen2019amass,
  title = {{AMASS}: Archive of Motion Capture as Surface Shapes},
  author = {Mahmood, Naureen and Ghorbani, Nima and Troje, Nikolaus F. and Pons-Moll, Gerard and Black, Michael J.},
  booktitle = {ICCV},
  pages = {5442--5451},
  year = {2019}
}

@inproceedings{xu2023eqmotion,
  title={Eqmotion: Equivariant multi-agent motion prediction with invariant interaction reasoning},
  author={Xu, Chenxin and Tan, Robby T and Tan, Yuhong and Chen, Siheng and Wang, Yu Guang and Wang, Xinchao and Wang, Yanfeng},
  booktitle={CVPR},
  pages={1410--1420},
  year={2023}
}

@inproceedings{andriluka2018posetrack,
  title={Posetrack: A benchmark for human pose estimation and tracking},
  author={Andriluka, Mykhaylo and Iqbal, Umar and Insafutdinov, Eldar and Pishchulin, Leonid and Milan, Anton and Gall, Juergen and Schiele, Bernt},
  booktitle={CVPR},
  pages={5167--5176},
  year={2018}
}

@article{gao2025social,
  title={Social-Pose: Enhancing Trajectory Prediction with Human Body Pose},
  author={Gao, Yang and Saadatnejad, Saeed and Alahi, Alexandre},
  journal={TITS},
volume={26},
  number={11},
  pages={21309--21319},
  year={2025}
}

@article{gao2024multi,
  title={Multi-transmotion: Pre-trained model for human motion prediction},
  author={Gao, Yang and Luan, Po-Chien and Alahi, Alexandre},
  journal={CoRL},
  year={2024}
}

@article{saadatnejad2023social,
  title={Social-transmotion: Promptable human trajectory prediction},
  author={Saadatnejad, Saeed and Gao, Yang and Messaoud, Kaouther and Alahi, Alexandre},
  journal={ICLR},
  year={2024}
}

@inproceedings{hosseininejad2025motionmap,
  title={MotionMap: Representing Multimodality in Human Pose Forecasting},
  author={Hosseininejad, Reyhaneh and Shukla, Megh and Saadatnejad, Saeed and Salzmann, Mathieu and Alahi, Alexandre},
  booktitle={CVPR},
  pages={22680--22689},
  year={2025}
}

@inproceedings{jiang2024worldpose,
  title={WorldPose: A world cup dataset for global {3D} human pose estimation},
  author={Jiang, Tianjian and Billingham, Johsan and M{\"u}ksch, Sebastian and Zarate, Juan and Evans, Nicolas and Oswald, Martin R and Polleyfeys, Marc and Hilliges, Otmar and Kaufmann, Manuel and Song, Jie},
  booktitle={ECCV},
  pages={343--362},
  year={2024}
}

@inproceedings{lee2019set,
  title={Set transformer: A framework for attention-based permutation-invariant neural networks},
  author={Lee, Juho and Lee, Yoonho and Kim, Jungtaek and Kosiorek, Adam and Choi, Seungjin and Teh, Yee Whye},
  booktitle={ICML},
  pages={3744--3753},
  year={2019}
}

@inproceedings{martinez2021pose,
  title={Pose transformers (potr): Human motion prediction with non-autoregressive transformers},
  author={Mart{\'\i}nez-Gonz{\'a}lez, Angel and Villamizar, Michael and Odobez, Jean-Marc},
  booktitle={ICCV},
  pages={2276--2284},
  year={2021}
}

@article{h36m_pami,
author = {Ionescu, Catalin and Papava, Dragos and Olaru, Vlad and Sminchisescu,  Cristian},
title = {Human3.6M: Large Scale Datasets and Predictive Methods for {3D} Human Sensing in Natural Environments},
journal = {TPAMI},
volume = {36},
number = {7},
pages = {1325-1339},
year = {2014}
}

@misc{cmu2003,
  title={CMU Graphics Lab Motion Capture Database},
  author={CMU Graphics Lab},
  howpublished={\url{http://mocap.cs.cmu.edu/}},
  year={2003}
}

@article{bouazizi2022motionmixer,
  title={Motionmixer: Mlp-based {3D} human body pose forecasting},
  author={Bouazizi, Arij and Holzbock, Adrian and Kressel, Ulrich and Dietmayer, Klaus and Belagiannis, Vasileios},
  journal={arXiv preprint arXiv:2207.00499},
  year={2022}
}

@inproceedings{chen2023humanmac,
  title={Humanmac: Masked motion completion for human motion prediction},
  author={Chen, Ling-Hao and Zhang, Jiawei and Li, Yewen and Pang, Yiren and Xia, Xiaobo and Liu, Tongliang},
  booktitle={ICCV},
  pages={9544--9555},
  year={2023}
}

@inproceedings{choudhury2023tempo,
  title={Tempo: Efficient multi-view pose estimation, tracking, and forecasting},
  author={Choudhury, Rohan and Kitani, Kris M and Jeni, L{\'a}szl{\'o} A},
  booktitle={ICCV},
  pages={14750--14760},
  year={2023}
}

@inproceedings{gao2023decompose,
  title={Decompose more and aggregate better: Two closer looks at frequency representation learning for human motion prediction},
  author={Gao, Xuehao and Du, Shaoyi and Wu, Yang and Yang, Yang},
  booktitle={CVPR},
  pages={6451--6460},
  year={2023}
}

@inproceedings{parsaeifard2021learning,
  title={Learning decoupled representations for human pose forecasting},
  author={Parsaeifard, Behnam and Saadatnejad, Saeed and Liu, Yuejiang and Mordan, Taylor and Alahi, Alexandre},
  booktitle={ICCV},
  pages={2294--2303},
  year={2021}
}

@inproceedings{jeong2024multi,
  title={Multi-agent long-term {3D} human pose forecasting via interaction-aware trajectory conditioning},
  author={Jeong, Jaewoo and Park, Daehee and Yoon, Kuk-Jin},
  booktitle={CVPR},
  pages={1617--1628},
  year={2024}
}

@inproceedings{rahman2023best,
  title={Best practices for 2-body pose forecasting},
  author={Rahman, Muhammad Rameez Ur and Scofano, Luca and De Matteis, Edoardo and Flaborea, Alessandro and Sampieri, Alessio and Galasso, Fabio},
  booktitle={CVPR},
  pages={3614--3624},
  year={2023}
}

@inproceedings{saadatnejad2023generic,
  title={A generic diffusion-based approach for {3D} human pose prediction in the wild},
  author={Saadatnejad, Saeed and Rasekh, Ali and Mofayezi, Mohammadreza and Medghalchi, Yasamin and Rajabzadeh, Sara and Mordan, Taylor and Alahi, Alexandre},
  booktitle={ICRA},
  pages={8246--8253},
  year={2023}
}

@inproceedings{xu2023auxiliary,
  title={Auxiliary tasks benefit {3D} skeleton-based human motion prediction},
  author={Xu, Chenxin and Tan, Robby T and Tan, Yuhong and Chen, Siheng and Wang, Xinchao and Wang, Yanfeng},
  booktitle={ICCV},
  pages={9509--9520},
  year={2023}
}

@inproceedings{adeli2021tripod,
  title={Tripod: Human trajectory and pose dynamics forecasting in the wild},
  author={Adeli, Vida and Ehsanpour, Mahsa and Reid, Ian and Niebles, Juan Carlos and Savarese, Silvio and Adeli, Ehsan and Rezatofighi, Hamid},
  booktitle={ICCV},
  pages={13390--13400},
  year={2021}
}

@inproceedings{cao2020long,
  title={Long-term human motion prediction with scene context},
  author={Cao, Zhe and Gao, Hang and Mangalam, Karttikeya and Cai, Qi-Zhi and Vo, Minh and Malik, Jitendra},
  booktitle={ECCV},
  pages={387--404},
  year={2020}
}

@inproceedings{tanke2023social,
  title={Social diffusion: Long-term multiple human motion anticipation},
  author={Tanke, Julian and Zhang, Linguang and Zhao, Amy and Tang, Chengcheng and Cai, Yujun and Wang, Lezi and Wu, Po-Chen and Gall, Juergen and Keskin, Cem},
  booktitle={ICCV},
  pages={9601--9611},
  year={2023}
}

@inproceedings{barquero2023belfusion,
  title={Belfusion: Latent diffusion for behavior-driven human motion prediction},
  author={Barquero, German and Escalera, Sergio and Palmero, Cristina},
  booktitle={ICCV},
  pages={2317--2327},
  year={2023}
}

@inproceedings{sun2024comusion,
  title={Comusion: Towards consistent stochastic human motion prediction via motion diffusion},
  author={Sun, Jiarui and Chowdhary, Girish},
  booktitle={ECCV},
  pages={18--36},
  year={2024}
}

@inproceedings{curreli2025nonisotropic,
  title={Nonisotropic gaussian diffusion for realistic {3D} human motion prediction},
  author={Curreli, Cecilia and Muhle, Dominik and Saroha, Abhishek and Ye, Zhenzhang and Marin, Riccardo and Cremers, Daniel},
  booktitle={CVPR},
  pages={1871--1882},
  year={2025}
}

@inproceedings{medjaouri2022hr,
  title={Hr-stan: High-resolution spatio-temporal attention network for {3D} human motion prediction},
  author={Medjaouri, Omar and Desai, Kevin},
  booktitle={CVPR},
  pages={2540--2549},
  year={2022}
}

@InProceedings{graber2020dynamic,
    author = {Graber, Colin and Schwing, Alexander G.},
    title = {Dynamic Neural Relational Inference},
    booktitle = {CVPR},
    month = {June},
    year = {2020}
}

@article{yao2024trajclip,
  title={TrajCLIP: Pedestrian trajectory prediction method using contrastive learning and idempotent networks},
  author={Yao, Pengfei and Zhu, Yinglong and Bi, Huikun and Mao, Tianlu and Wang, Zhaoqi},
  journal={NeurIPS},
  volume={37},
  pages={77023--77037},
  year={2024}
}

@inproceedings{bae2024singulartrajectory,
  title={Singulartrajectory: Universal trajectory predictor using diffusion model},
  author={Bae, Inhwan and Park, Young-Jae and Jeon, Hae-Gon},
  booktitle={CVPR},
  pages={17890--17901},
  year={2024}
}

@inproceedings{fang2025neuralized,
  title={Neuralized Markov Random Field for Interaction-Aware Stochastic Human Trajectory Prediction.},
  author={Fang, Zilin and Hsu, David and Lee, Gim Hee},
  booktitle={ICLR},
  year={2025}
}

@inproceedings{aydemir2023adapt,
  title={Adapt: Efficient multi-agent trajectory prediction with adaptation},
  author={Aydemir, G{\"o}rkay and Akan, Adil Kaan and G{\"u}ney, Fatma},
  booktitle={ICCV},
  pages={8295--8305},
  year={2023}
}

@inproceedings{choi2023r,
  title={R-pred: Two-stage motion prediction via tube-query attention-based trajectory refinement},
  author={Choi, Sehwan and Kim, Jungho and Yun, Junyong and Choi, Jun Won},
  booktitle={ICCV},
  pages={8525--8535},
  year={2023}
}

@inproceedings{gu2021densetnt,
  title={Densetnt: End-to-end trajectory prediction from dense goal sets},
  author={Gu, Junru and Sun, Chen and Zhao, Hang},
  booktitle={ICCV},
  pages={15303--15312},
  year={2021}
}

@inproceedings{jiang2023motiondiffuser,
  title={Motiondiffuser: Controllable multi-agent motion prediction using diffusion},
  author={Jiang, Chiyu and Cornman, Andre and Park, Cheolho and Sapp, Benjamin and Zhou, Yin and Anguelov, Dragomir and others},
  booktitle={CVPR},
  pages={9644--9653},
  year={2023}
}

@inproceedings{li2022graph,
  title={Graph-based spatial transformer with memory replay for multi-future pedestrian trajectory prediction},
  author={Li, Lihuan and Pagnucco, Maurice and Song, Yang},
  booktitle={CVPR},
  pages={2231--2241},
  year={2022}
}

@inproceedings{rowe2023fjmp,
  title={Fjmp: Factorized joint multi-agent motion prediction over learned directed acyclic interaction graphs},
  author={Rowe, Luke and Ethier, Martin and Dykhne, Eli-Henry and Czarnecki, Krzysztof},
  booktitle={CVPR},
  pages={13745--13755},
  year={2023}
}

@inproceedings{capellera2025jointdiff,
  title={JointDiff: Bridging Continuous and Discrete in Multi-Agent Trajectory Generation},
  author={Capellera, Guillem and Ferraz, Luis and Rubio, Antonio and Alahi, Alexandre and Agudo, Antonio},
  booktitle={ICLR},
  year={2026}
}

@inproceedings{capellera2025unified,
  title={Unified uncertainty-aware diffusion for multi-agent trajectory modeling},
  author={Capellera, Guillem and Rubio, Antonio and Ferraz, Luis and Agudo, Antonio},
  booktitle={CVPR},
  pages={22476--22486},
  year={2025}
}

@inproceedings{capellera2024transportmer,
  title={Transportmer: A holistic approach to trajectory understanding in multi-agent sports},
  author={Capellera, Guillem and Ferraz, Luis and Rubio, Antonio and Agudo, Antonio and Moreno-Noguer, Francesc},
  booktitle={ACCV},
  pages={1652--1670},
  year={2024}
}

@article{gutierrez2026once,
  title={Once Upon a Goal: Towards orientation-based shot metrics in football},
  author={Guti{\'e}rrez-P{\'e}rez, Marc and Yeung, Calvin and Fujii, Keisuke and Agudo, Antonio},
  journal={CVIU},
  pages={104732},
  year={2026}
}

@article{gutierrez4998149pnlcalib,
  title={Pnlcalib: Sports Field Registration Via Points and Lines Optimization},
  author={Guti{\'e}rrez-P{\'e}rez, Marc and Agudo, Antonio},
  journal={CVIU},
  pages={104712},
  year={2026}

}

@inproceedings{gutiACCV26,
  title={ Learning Human Sports Motion Feasibility via Quantile Regression on Tracking Data},
  author={ Guti{\'e}rrez-P{\'e}rez, Marc and Agudo, Antonio},
  booktitle={ACCV},
  year={2026}
}

@article{teoh2025coherent,
  title={Coherent Multi-Agent Trajectory Forecasting in Team Sports with CausalTraj},
  author={Teoh, Wei Zhen},
  journal={AAAIW},
  year={2026}
}

@inproceedings{zaier2023dual,
  title={A dual perspective of human motion analysis-{3D} pose estimation and {2D} trajectory prediction},
  author={Zaier, Mayssa and Wannous, Hazem and Drira, Hassen and Boonaert, Jacques},
  booktitle={ICCV},
  pages={2189--2199},
  year={2023}
}

@inproceedings{zheng2025efficient,
  title={Efficient Multi-Person Motion Prediction by Lightweight Spatial and Temporal Interactions},
  author={Zheng, Yuanhong and Yu, Ruixuan and Sun, Jian},
  booktitle={ICCV},
  pages={10844--10853},
  year={2025}
}

@article{capelleraTPAMI26,
  title={Heteroscedastic Diffusion for Multi-Agent Trajectory Modeling},
  author={Capellera, Guillem and Rubio, Antonio and Ferraz, Luis and Agudo, Antonio},
  journal={TPAMI},
  volume={48},
  number={10},
  pages={11964--11977},
  year={2026}
}

@article{wang2021multi,
  title={Multi-person {3D} motion prediction with multi-range transformers},
  author={Wang, Jiashun and Xu, Huazhe and Narasimhan, Medhini and Wang, Xiaolong},
  journal={NeurIPS},
  volume={34},
  pages={6036--6049},
  year={2021}
}

@inproceedings{peng2023trajectory,
  title={Trajectory-aware body interaction transformer for multi-person pose forecasting},
  author={Peng, Xiaogang and Mao, Siyuan and Wu, Zizhao},
  booktitle={CVPR},
  pages={17121--17130},
  year={2023}
}

@article{vendrow2022somoformer,
  title={Somoformer: Multi-person pose forecasting with transformers},
  author={Vendrow, Edward and Kumar, Satyajit and Adeli, Ehsan and Rezatofighi, Hamid},
  journal={arXiv preprint arXiv:2208.14023},
  year={2022}
}

@article{gao2025omnitraj,
  title={Omnitraj: Pre-training on heterogeneous data for adaptive and zero-shot human trajectory prediction},
  author={Gao, Yang and Luan, Po-Chien and Messaoud, Kaouther and Feng, Lan and Alahi, Alexandre},
  journal={arXiv preprint arXiv:2507.23657},
  year={2025}
}

@inproceedings{peral2026multi,
  title={Multi-modal soccer scene analysis with masked pre-training},
  author={Peral, Marc and Capellera, Guillem and Ferraz, Luis and Rubio, Antonio and Agudo, Antonio},
  booktitle={WACV},
  pages={3536--3545},
  year={2026}
}

@inproceedings{peral2026synergy,
  title={Synergy Between Trajectories And Human Pose for Soccer},
  author={Peral, Marc and Capellera, Guillem and Ferraz, Luis and Rubio, Antonio and Gra{\ss}hof, Stella and Hansen, Dan Witzner and Agudo, Antonio},
  booktitle={ICIP},
  year={2026}
}

@inproceedings{capellera2024footbots,
  title={Footbots: A transformer-based architecture for motion prediction in soccer},
  author={Capellera, Guillem and Ferraz, Luis and Rubio, Antonio and Agudo, Antonio and Moreno-Noguer, Francesc},
  booktitle={ICIP},
  pages={2313--2319},
  year={2024}
}
}
\newpage

\setcounter{page}{1}
\setcounter{section}{0}
\renewcommand{\thesection}{\Alph{section}}
\setcounter{figure}{0}
\renewcommand{\thefigure}{A\arabic{figure}}
\setcounter{table}{0}
\renewcommand{\thetable}{A\arabic{table}}

\twocolumn[{%
\renewcommand\twocolumn[1][]{#1}%
\maketitlesupp
}]

\section{Joint Balancing Strategy}
\begin{figure}[t]
    \centering
    \includegraphics[width=0.8\linewidth]{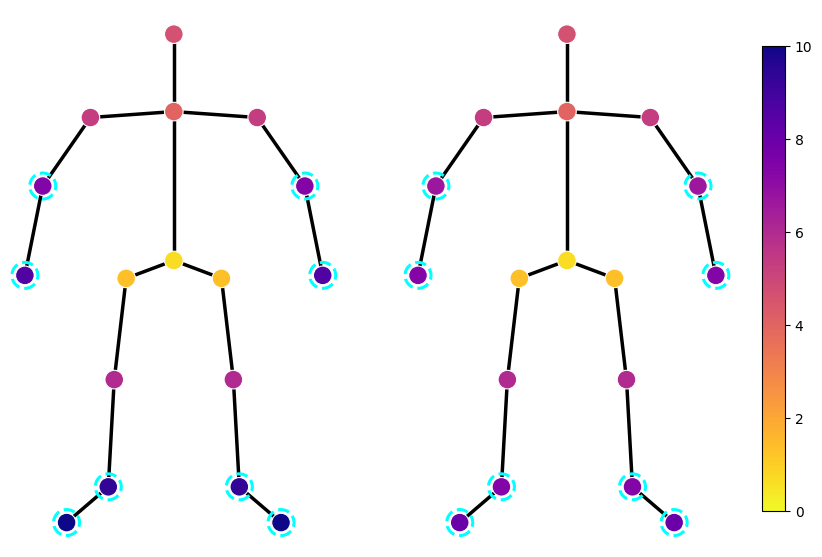}
    \caption{\textbf{Per-joint error under uniform loss weighting (left) and with joint rebalancing (right).} Each joint is colored by its mean absolute error on the test set, on a shared scale from $0$ (light, low error) to $10$ (dark, high error) across both skeletons, so the two panels are directly comparable. The targeted distal joints (elbows, wrists, ankles, and feet), outlined in cyan, are those with consistently above-average error under uniform weighting, and rebalancing visibly reduces their error.
    }
    \label{joints-poses}
\end{figure}

The target joints are identified by directly measuring per-joint error under uniform weighting and selecting those with consistently above-average error, rather than by assumption. Under uniform weighting the distal joints (elbows, wrists, ankles, and feet) carry markedly higher error than the rest of the skeleton, while proximal joints such as the hips, spine, and shoulders remain comparatively stable in root-relative coordinates and dominate the averaged loss. Figure~\ref{joints-poses} visualizes per-joint error on a shared color scale before (left) and after (right) rebalancing, so the two skeletons are directly comparable. The reduction is concentrated in the targeted distal joints, whose error visibly drops after rebalancing. This directly prevents the degenerate static solution into which a uniformly weighted loss could settle if the model minimizes average error by collapsing toward a near-motionless mean pose and leaving exactly these dynamic joints poorly predicted. By steering optimization toward the joints that carry the most discriminative information about player actions, the rebalancing lowers their error and incentivizes fine-grained articulation instead of collapse.

\section{Trajectory Smoothing Exploration}
The frequency-domain smoothing term was the result of an extended exploration of alternative strategies for suppressing the high-frequency oscillation that appears in absolute pitch-coordinate trajectory predictions. The root cause of this jitter is that the model predicts absolute positions at each frame independently, with no explicit constraint on temporal consistency between consecutive predictions.

The most natural alternative is to predict frame-to-frame displacements rather than absolute positions, converting the entire input sequence to a displacement representation so that the model reasons about changes in position rather than positions themselves. This was implemented but produced a significant degradation of results. When all positions are expressed as displacements, the model loses access to the absolute spatial context of the scene: it can no longer directly represent where each player is on the pitch relative to others, which is precisely the information the social attention block relies on to model inter-player interactions.

A second variant was then explored, where the observable part of the sequence was kept as absolute positions and only the frames to be forecast were expressed as displacements. This reduced jitter and improved displacement error in the first forecast frames, but increased general ADE and specially FDE as errors accumulated over the displacement chain. Moreover, this representation fixed the boundary between the observed and forecast portions of the sequence at a specific frame index, removing the model's flexibility to operate with different observation and forecast horizons.

A third alternative was a local smoothness penalty computed in the time domain, penalizing second-order differences in position between consecutive frame triplets, which correspond to sudden changes in velocity. This formulation operates only on pairs of adjacent frames and therefore misses oscillations that span more than two frames, which are the dominant artifact in practice.

None of these alternatives matched the effectiveness of the frequency-domain formulation ultimately adopted, which penalizes the full-sequence power spectrum weighted by frequency, capturing all systematic high-frequency oscillations regardless of their temporal scale within the forecasting horizon.

\end{document}